# Are Arabic Benchmarks Reliable? QIMMA's Quality-First Approach to LLM Evaluation

**Leen AlQadi, Ahmed Alzubaidi, Mohammed Alyafeai, Hamza Alobeidli, Maitha Alhammadi, Shaikha Alsuwaidi, Omar Alkaabi, Basma El Amel Boussaha, Hakim Hacid**
Technology Innovation Institute, Abu Dhabi, UAE
leen.alqadi@tii.ae

## Abstract

We present QIMMA, a quality-assured Arabic LLM leaderboard that places systematic benchmark validation at its core. Rather than aggregating existing resources as-is, QIMMA applies a multi-model assessment pipeline combining automated LLM judgment with human review to surface and resolve systematic quality issues in well-established Arabic benchmarks before evaluation. The result is a curated, multi-domain, multi-task evaluation suite of over 52k samples, grounded predominantly in native Arabic content; code evaluation tasks are the sole exception, as they are inherently language-agnostic. Transparent implementation via LightEval, EvalPlus and public release of per-sample inference outputs make QIMMA a reproducible and community-extensible foundation for Arabic NLP evaluation.

## 1 Introduction

The rapid advancement of Large Language Models has transformed Natural Language Processing (NLP), spurring a growing body of Arabic NLP evaluation efforts that reflect the community's commitment to serving over 400 million speakers across diverse dialects and cultural contexts (Al-Khalifa et al., 2025). While native Arabic resources have emerged across a range of domains, they vary considerably in quality, coverage, and validation rigor, leaving the evaluation landscape without a unified, quality-assured foundation.

Despite this progress, existing Arabic benchmarks share limitations that collectively undermine their reliability. Many rely on translation from English sources, which prior work has shown to introduce distributional shifts and evaluation artifacts in cross-lingual settings (Artetxe et al., 2020). Even native resources are often released without rigorous quality validation, as evidenced by recent efforts that uncover annotation inconsistencies and cultural misalignment in established benchmarks (Nacar et al., 2025). Coverage remains fragmented across isolated tasks and narrow domains, and evaluation scripts and per-sample outputs are not consistently released publicly, limiting reproducibility and community-driven improvement (Almatham et al., 2025; Alzubaidi et al., 2025).

We present QIMMA قِمّة (Arabic for "summit"), a quality-assured Arabic LLM leaderboard built on native content curation and systematic quality validation. QIMMA consolidates 109 subsets spanning cultural, STEM, legal, medical, and other domains, into a unified multi-task framework, spanning multiple-choice, generative, and code evaluation, comprising over 52k samples. Over 99% of content is native Arabic; the sole exception is code evaluation. At the heart of QIMMA is a multi-model validation pipeline that identifies and resolves quality issues in source benchmarks before they reach evaluation, ensuring reported scores reflect genuine model capability rather than benchmark artifacts.

This paper contributes the following:

- **Quality validation pipeline:** A methodology combining multi-model automated assessment with human review; a reusable approach that can be applied beyond QIMMA to audit and refine any Arabic evaluation resource.
- **Empirical findings:** Through systematic application of this pipeline, evidence of recurring quality issues across well-established Arabic benchmarks are identified, demonstrating that unvalidated resources pose concrete risks to evaluation validity.
- **QIMMA leaderboard:** Introduction of a unified, multi-domain, multi-task Arabic LLM evaluation framework grounded in native Arabic content, with fully transparent implementation[1].

[1]Leaderboard available on https://huggingface.co

The remainder of this paper is organized as follows: Section 2 positions QIMMA within the broader Arabic NLP evaluation landscape, Section 3 describes the leaderboard construction (covering dataset selection, the quality validation pipeline, and evaluation implementation), Section 4 presents our systematic examination of quality issues identified across benchmarks, Section 5 reports evaluated models' results and findings, and Section 6 concludes.

## 2 Related Work

The growing importance of Arabic in the global NLP landscape has spurred substantial investment in evaluation resources, yet the field remains without a unified, quality-assured foundation. This section surveys existing benchmarks, leaderboards, and quality validation efforts, situating QIMMA within this landscape.

### 2.1 Arabic NLP Evaluation Landscape

Recent surveys document rapid growth in Arabic LLM evaluation. Alzubaidi et al. (2025) provide the first systematic review of 40+ Arabic benchmarks, proposing a taxonomy spanning Knowledge and STEM, NLP Tasks, Culture and Dialects, and Target-Specific evaluations, while identifying cultural misalignment in translated datasets as a recurring concern. Al-Khalifa et al. (2025) trace the historical evolution of Arabic NLP and current research trends, and Mashaabi et al. (2024) analyze pretraining data and dialectal coverage across existing resources.

Individual benchmark efforts have grown considerably in domain breadth. Koto et al. (2024) introduced ArabicMMLU, a large-scale native Arabic MCQ benchmark later extended to dialectal Arabic by Altakrori et al. (2025). Domain-specific efforts address legal reasoning (Abu Shairah et al., 2025), poetry understanding (Alghallabi et al., 2025), and medical knowledge (Daoud et al., 2025). For broader coverage, LAraBench (Abdelali et al., 2024) spans dozens of tasks, though it acknowledges limitations in cultural alignment and translation dependence.

### 2.2 Arabic LLM Leaderboards

Centralized leaderboards have emerged as the primary mechanism for tracking Arabic LLM progress. OALL (El Filali et al., 2024) pioneered open-source rankings, with OALL v2 (El Filali et al., 2025) addressing benchmark saturation by transitioning to predominantly native content, including ArabicMMLU, ALRAGE, and AraTrust though retaining some translated material and lacking quality validation. ILMAAM (Nacar et al., 2025) takes a culturally aligned approach, with eleven experts annotating over 2,500 questions to filter culturally misaligned content, but provides only partial data access. SILMA.AI's Arabic Broad Leaderboard (SILMA.AI, 2024) incorporates human validation and public inference results but remains limited in scale. HELM Arabic (CRFM, 2023) provides standardized and reproducible evaluation protocols, while AraGen (AI and MBZUAI, 2024) introduces dynamic generative evaluation addressing limitations of static MCQ benchmarks. BALSAM (Almatham et al., 2025) offers comprehensive evaluation across 78 tasks with private test sets to prevent contamination, but covers only around 50% native Arabic content and lacks quality validation. Table 1 compares these platforms across key dimensions.

| Leaderboard | Open Source | Native Arabic | Quality Valid. | Code Eval | Public Outputs |
|---|---|---|---|---|---|
| OALL v1 | ✓ | Mixed | × | × | ✓ |
| OALL v2 | ✓ | Mostly | × | × | ✓ |
| BALSAM | Partial | 50% | × | × | × |
| AraGen | ✓ | 100% | × | × | × |
| SILMA ABL | ✓ | 100% | ✓ | × | ✓ |
| ILMAAM | Partial | 100% | ✓ | × | × |
| HELM Arabic | ✓ | Mixed | × | × | ✓ |
| **QIMMA** | ✓ | **99%** | ✓ | ✓ | ✓ |

Table 1: Comparison of Arabic LLM Leaderboards. QIMMA is the only platform combining open source, quality validation, code evaluation, and public inference results.

### 2.3 Benchmark Quality and Validation

Research across NLP broadly has documented systematic quality problems that compromise evaluation validity, from annotation artifacts that allow models to exploit spurious patterns (Gururangan et al., 2018), to adversarial findings revealing benchmark brittleness (Nie et al., 2020; Kiela et al., 2021), to arguments that static single-metric benchmarks no longer suffice for rigorous model assessment (Ruder, 2021). For Arabic specifically, Alzubaidi et al. (2025) identify translation artifacts, inconsistent methodologies, and limited reproducibility as recurring concerns, while Nacar et al. (2025) highlight cultural bias in ground truth labels, noting that benchmarks often embed region-specific per-

/spaces/qimma/leaderboard

spectives as universal truth. Despite this awareness, systematic quality validation combining automated and human review remains rare in Arabic benchmark development.

QIMMA directly addresses this gap by applying a multi-model validation pipeline across all included benchmarks before evaluation. Where existing work provides either comprehensive coverage (BALSAM) or specialized assessment (ArabicMMLU, ALARB), QIMMA combines both while emphasizing quality validation and transparency.

## 3 QIMMA Leaderboard

### 3.1 Dataset Selection and Coverage

We conducted a systematic survey of available Arabic NLP datasets to identify candidates for inclusion in QIMMA. Our selection criteria prioritized: (1) **native Arabic content** rather than translations to preserve linguistic authenticity and cultural nuance, (2) **domain diversity** to enable a well rounded model assessment, (3) **task type variety** covering multiple-choice, generative, and code evaluation, and (4) **coverage of linguistic variation** including both Modern Standard Arabic (MSA) and dialectal variants where appropriate.

This process identified 14 datasets organized into 109 distinct subsets. Table 2 presents our dataset inventory organized by domain. Approximately 99% of our content consists of native Arabic data, with the exception of code evaluation tasks that are inherently language-agnostic. This native content emphasis distinguishes QIMMA from benchmarks that rely heavily on translation, which inevitably loses cultural context and linguistic nuance critical for authentic evaluation. Our coverage spans seven domains:

**Cultural Understanding** is covered by AraDiCE-Culture (Mousi et al., 2025), ArabCulture (Sadallah and Koto, 2025), and PalmX (Alwajih et al., 2025), collectively probing regional knowledge across the Gulf, Levant, Egypt, and North Africa.

**STEM** is addressed through three complementary benchmarks spanning scientific knowledge, quantitative reasoning, and verbal aptitude. ArabicMMLU (Koto et al., 2024) draws from school exams across eight Arab countries, covering STEM subjects, social sciences, humanities, and Arabic language. The 3LM STEM benchmarks (Boussaha et al., 2025) provide native and synthetically generated questions in physics, chemistry, biology, mathematics, and geography. GAT (AlBallaa et al., 2025), derived from the Saudi General Aptitude Test, assesses mathematics, verbal reasoning, and language comprehension; skills that reflect the quantitative and analytical aptitude central to STEM education and evaluation.

**Legal** reasoning is assessed at two levels of jurisdiction: ArabLegalQA (Hijazi et al., 2024) evaluates Saudi legal knowledge, while MizanQA (Bahaj and Ghogho, 2025) targets Moroccan law, together capturing the diversity of Arabic legal systems.

**Poetry and Literature** is addressed by FannOrFlop (Alghallabi et al., 2025), which evaluates classical and modern Arabic poetry comprehension and literary analysis.

**Medical** knowledge is covered by two benchmarks: MedArabiQ (Daoud et al., 2025) and MedAraBench (Daoud et al., 2026), which together probe healthcare terminology and clinical reasoning across MCQ and open-ended QA formats.

**Trust & Safety** is evaluated through AraTrust (Alghamdi et al., 2024), covering truthfulness, ethics, privacy, and safety dimensions.

**Code Generation** is evaluated through the 3LM Arabic adaptations of HumanEval+ and MBPP+ (Boussaha et al., 2025), making QIMMA the first Arabic leaderboard to include code generation in its evaluation suite.

### 3.2 Quality Validation Pipeline

Benchmark reliability depends critically on sample quality, and errors in gold answers, cultural bias, or malformed questions directly compromise evaluation validity. To address this, we developed a multi-stage validation pipeline combining automated multi-model assessment with human review. The pipeline overview is shown in Figure 1.

#### 3.2.1 Multi-Model Automated Assessment

The automated assessment of each sample capitalized on the capabilities of two state-of-the-art LLMs: `Qwen3-235B-A22B-Instruct` (Yang et al., 2025) and `DeepSeek-V3-671B` (DeepSeek-AI et al., 2025). Model selection was guided by three criteria: strong Arabic language capability, reliable instruction-following on long structured prompts, and open accessibility to support reproducibility. Both models satisfy these criteria

| Benchmark | Task Type | Subsets | Domain | Quality Filter | |
|---|---|---|---|---|---|
| | | | | Kept | Discarded |
| AraDiCE-Culture | MCQ | 6 | Cultural Understanding | 180 | 0 |
| ArabCulture | MCQ | 13 | Cultural Commonsense | 3,475 | 7 |
| PalmX | MCQ | 2 | Islamic Studies, Cultural Understanding | 2,976 | 25 |
| ArabicMMLU | MCQ | 40 | STEM, Social Sciences, Humanities, Arabic Language | 13,727 | 436 |
| GAT | MCQ | 9 | Maths, Verbal Reasoning, Language Comprehension | 13,985 | 1 |
| 3LM STEM | MCQ | 2 | STEM (Physics, Chemistry, Biology, Maths, Geography) | 2,608 | 1 |
| ArabLegalQA | QA | 1 | Legal Reasoning (Saudi Law) | 79 | 0 |
| MizanQA | MCQ | 1 | Legal Reasoning (Moroccan Law) | 1,728 | 41 |
| MedArabiQ | MCQ | 3 | Medical | 299 | 1 |
| | QA | 2 | | 200 | 0 |
| MedAraBench | MCQ | 26 | | 4,927 | 33 |
| FannOrFlop | QA | 1 | Poetry & Literary Analysis | 6,941 | 43 |
| AraTrust | MCQ | 1 | Trust & Safety | 522 | 0 |

(a) Dataset inventory with quality filter results (kept/discarded counts).

| Benchmark | Total | Modified | Unchanged | Modification Rate |
|---|---|---|---|---|
| 3LM HumanEval+ | 164 | 145 | 19 | 88% |
| 3LM MBPP+ | 378 | 308 | 70 | 81% |

(b) 3LM Code benchmark Arabic prompt modification statistics.

Table 2: QIMMA dataset inventory and quality refinement statistics across all benchmarks. 3LM contributes both a STEM and a Code benchmark, accounting for 14 source datasets in total.

and differ in training data compositions, making their combined assessment more robust than either model alone.

Each model evaluates the benchmark samples against explicit quality criteria, and a binary score (0 or 1) is awarded for each passed criterion. There are ten validation points in the assessment rubric, awarding a binary score (0 or 1) per criterion (see Figure 1); the full prompt templates used for assessment are provided in Appendix C.

Samples are flagged for human review if either model scores below threshold (<7/10) or if the two models disagree significantly, leveraging both the scale of automated assessment and the bias reduction of model diversity.

#### 3.2.2 Human Annotation and Review

Samples flagged by automated assessment are reviewed manually by native Arabic speakers familiar with the cultural and dialectal nuances of the language. Annotators made final judgments on cultural context, dialectal variation, subjective interpretation, and subtle quality issues that automated assessment may have missed. For culturally sensitive content, diverse perspectives were considered, as correctness may vary across cultural contexts.

### 3.3 Evaluation Implementation

Following quality validation, we built a unified evaluation pipeline to assess SOTA Arabic LLMs across the curated benchmark suite. The following subsections detail the framework, prompting strategy, and metrics used.

#### 3.3.1 Evaluation Framework

QIMMA's evaluation relies primarily on LightEval framework (Habib et al., 2023), which unifies the majority of benchmarks under a single evaluation codebase. This choice was motivated by three considerations: the included benchmarks varied widely in what they released alongside their datasets, from complete evaluation pipelines to dataset files only, making a unified framework necessary for consistency; LightEval is well-established in the multilingual evaluation community and has been adopted by prior leaderboard initiatives (El Filali et al., 2024, 2025); and its standardized protocols support the transparency and reproducibility commitments central to QIMMA[2].

#### 3.3.2 Prompt Functions and Metrics

The system prompting strategy was standardized according to question format: (MCQ, QA, or multi-select) and whether the task required additional contextual information. The full set of prompt templates is illustrated in Figure 4 (Appendix B). Templates were assigned by format: MCQ for AraDiCE-Culture, AraTrust, MedArabiQ, 3LM STEM, PalmX, and GAT; MCQ-C for GAT (Reading) and ArabicMMLU subsets requiring passage comprehension; MCQ-I for GAT analogy, completion, and contextual subtasks; QA and QA-F for MedArabiQ; and QA-C for ArabLegalQA. For MizanQA and ArabCulture, we adopted the exact system prompts reported in the original works, as they include benchmark-specific contextual details.

[2]Code and datasets available on `https://github.com/tiiuae/QIMMA-leaderboard`

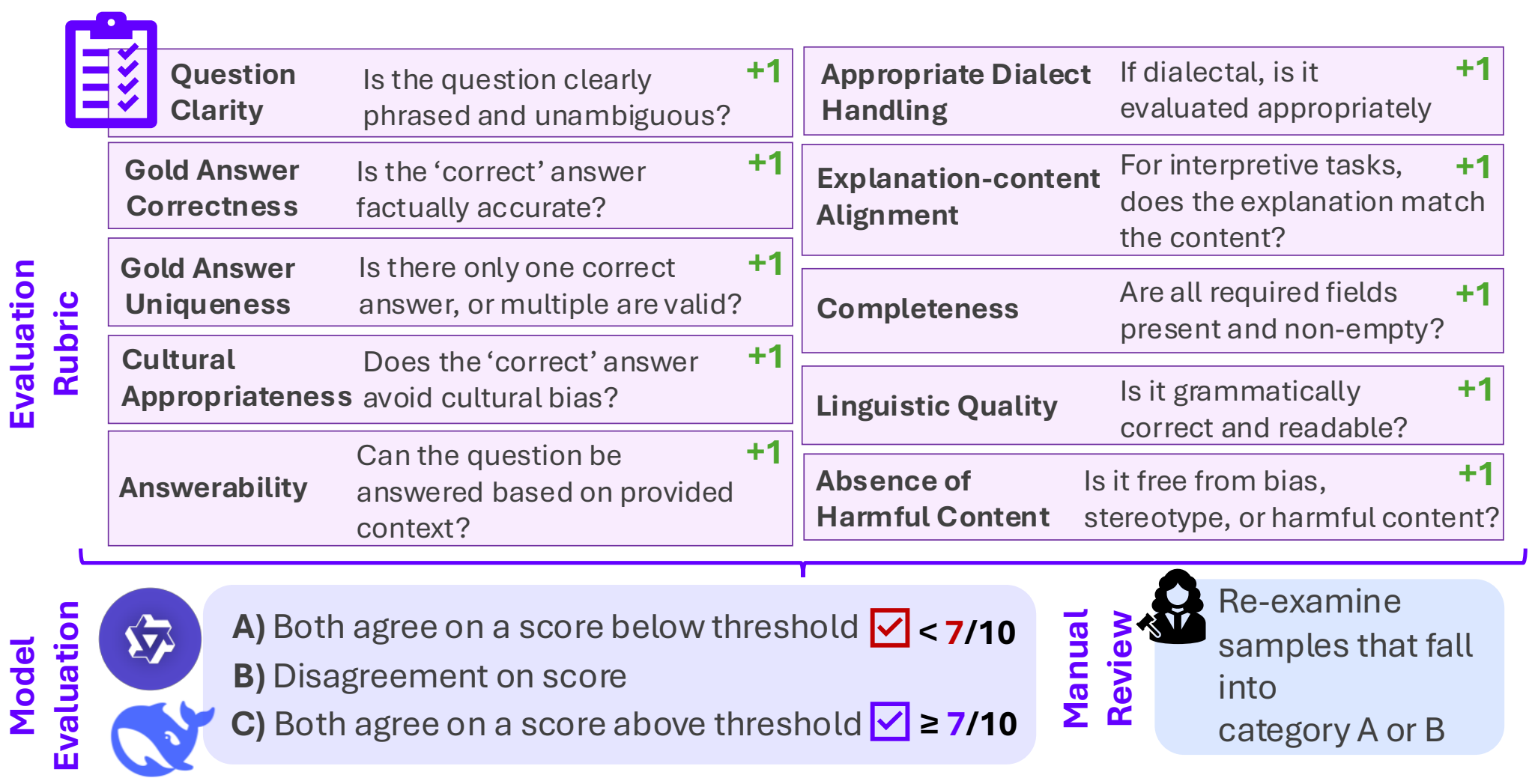

Figure 1: Overview of the multi-stage quality validation pipeline.

The evaluation metric was similarly determined by question type. For each MCQ instance, we denote the question as $q$ and the candidate options as $C = [c_1, c_2, \ldots, c_n]$, where each $c_i$ corresponds to a labeled choice (أ,ب,ج...), with correct answer index $g$. For QA, each instance consists of question $q$ and reference answer $g$.

- **Pass@1.** For code generation tasks (HumanEval+ and MBPP+), functional correctness is measured via pass@1, defined as the fraction of problems solved by a single generated solution.

- **Accuracy.** For MCQ benchmarks, accuracy is computed via Normalized Log-Likelihood (Eq. 1), with prediction inferred by Eq. 2. A prediction is correct when $p = g$. Applied to AraDiCE-Culture, AraTrust, ArabicMMLU, ArabCulture, 3LM STEM, PalmX, MedArabiQ, GAT, and MedAraBench.

$$l_i = \log P(c_i|q) \tag{1}$$

$$p = \arg\max_i \ L_i \tag{2}$$

- **Probability.** For multi-select cases (MizanQA), we return the probability mass assigned to the gold choices. Log-probabilities are computed via Eq. 1, normalized via softmax (Eq. 3), and aggregated over the gold index set $G$ (Eq. 4).

$$\hat{P}(c_i|q) = softmax(l_i) = \frac{\exp(l_i)}{\sum_{j=0}^{n} exp(l_j)} \tag{3}$$

$$p_G = \sum_{i \in G} \hat{P}(c_i|q) \tag{4}$$

- **BERTScore.** For generative tasks, semantic similarity between generated response $r$ and reference $g$ is measured using F1 BERTScore, computed with `aubmindlab/bert-base-arabertv02` (Antoun et al., 2020) for its specialization in Arabic text. Applied to MedArabiQ, ArabLegalQA, and FannOrFlop.

QIMMA provides a web-based interface with filters by domain, task type, and linguistic register (MSA vs. dialectal). Performance visualizations enable easy comparison across models and subsets. Critically, we make all model inference answers publicly accessible, not just aggregate scores, but detailed error analysis, quality auditing, and community-driven improvement. This transparency distinguishes QIMMA from closed evaluation systems and supports scientific reproducibility.

## 4 Quality Analysis and Findings

Our quality validation pipeline revealed systematic issues across existing Arabic benchmarks. This section presents a taxonomy of identified issues and their implications for evaluation validity.

### 4.1 MCQ and QA Benchmarks: Identified Issues

Our multi-model assessment and human review revealed recurring quality issues across the evaluated benchmarks. Rather than isolated errors, these patterns reflect systematic gaps in how existing Arabic benchmarks were constructed and validated. Figure 2 summarizes the discard rate across benchmarks with issues detected; Figure 3 (Appendix B) further shows that scale alone is neither sufficient nor necessary to predict quality, with benchmark-specific construction practices emerging as the more determinative factor. Notably, discard rates within ArabicMMLU vary substantially across subsets (Figure 6, Appendix B), with certain subjects exceeding 80%, likely reflecting differences in the preprocessing quality of underlying source materials rather than any property of the subject matter itself. The following discussion highlights benchmarks where failure modes were most pronounced; remaining benchmarks exhibited lower-impact instances of similar issue types.

**Answer quality** emerged as the most consequential category of issues. A recurring problem was incorrect or missing gold answers, which took several forms: factually wrong answers marked as correct, gold indices pointing to non-existent options, answer fields storing raw text that does not match any of the listed choices, and entirely empty or null answer fields. In ArabicMMLU, some samples contained no correct option among the listed choices, rendering the question unanswerable. Some instances in PalmX and MedAraBench had null or missing answer fields or gold answers that did not correspond to any listed choice.

**Question and formatting quality** issues were widespread across benchmarks. The most severe cases involved corrupt or illegible text caused by encoding errors, resulting in questions with garbled or repeated characters that were unreadable. Beyond encoding corruption, spelling and grammar errors were also identified, most notably in ArabicMMLU. Another distinct issue appeared where questions referred to external material, such as tables, absent from the sample entirely, making the question unanswerable regardless of the model's knowledge. Duplicate samples were identified across multiple benchmarks, most notably in MizanQA, and required deduplication to prevent inflated or misleading evaluation results.

**Cultural issues and stereotypes** represent a subtler but equally important category. Some cultural benchmarks conflated stereotype reinforcement with legitimate cultural knowledge, framing essentialist generalizations about specific populations as objectively correct answers. This was most evident in ArabCulture, where phrases treating diverse communities as monolithic (e.g. attributing uniform behaviors or values to an entire national or gender group) reflect a fundamental question design flaw: such questions assess whether a model has internalized societal biases rather than testing genuine cultural competence.

**Gold answer compliance in QA benchmarks** raises a principled concern about evaluation consistency: when a prompt instructs the model to follow strict output guidelines, gold answers must have been held to those same standards. FannOrFlop provides the clearest illustration of this, where discarded samples violated the benchmark's own structured output protocol in the gold answers themselves, through misaligned verse-explanation pairs, weak or non-verse-specific explanations, nonsensical or incoherent interpretations, empty or placeholder fields, explanations referencing content absent from the verse, irrelevant additions, linguistic anomalies such as unsolicited grammatical analysis, and formatting inconsistencies. Whether structural or content-related, these failures share a common root: reference answers were not held to the same standards the evaluation protocol demands of model outputs, rendering scoring against them invalid.

### 4.2 Code Benchmarks: Prompt Refinement

Code evaluation benchmarks required a fundamentally different quality intervention than MCQ and QA benchmarks. Rather than removing problematic samples, our quality work focused on refining the Arabic problem statements in 3LM Code Arabic adaptations of HumanEval+ and MBPP+ (Boussaha et al., 2025)[3], while leaving task identifiers, reference solutions, and evaluation test suites entirely unchanged.

Across both benchmarks, modifications fell into five recurring categories. **Linguistic refinement** in-volved normalization toward more natural Modern Standard Arabic and consistent imperative style, for example, standardizing verb choices such as أعد → أرجع to produce more idiomatic instruction phrasing. **Clarity improvements** addressed ambiguous instructions, unclear constraint phrasing, and examples that were difficult to interpret without additional context. **Consistency normalization** standardized mathematical and programming terminology, punctuation, quoting style, and example formatting across prompts within the same benchmark. **Structural corrections** fixed prompt formatting issues that could affect parsing or readability, such as broken triple-quoted strings, indentation errors, and corrupted merged text fragments, alongside restoration of any inadvertently altered fields. Finally, **semantic refinements** involved small wording adjustments to improve mathematical precision, for instance, clarifying whether a range is inclusive or exclusive, while preserving the intended task behavior.

[3] https://huggingface.co/datasets/tiiuae/evalplus-arabic

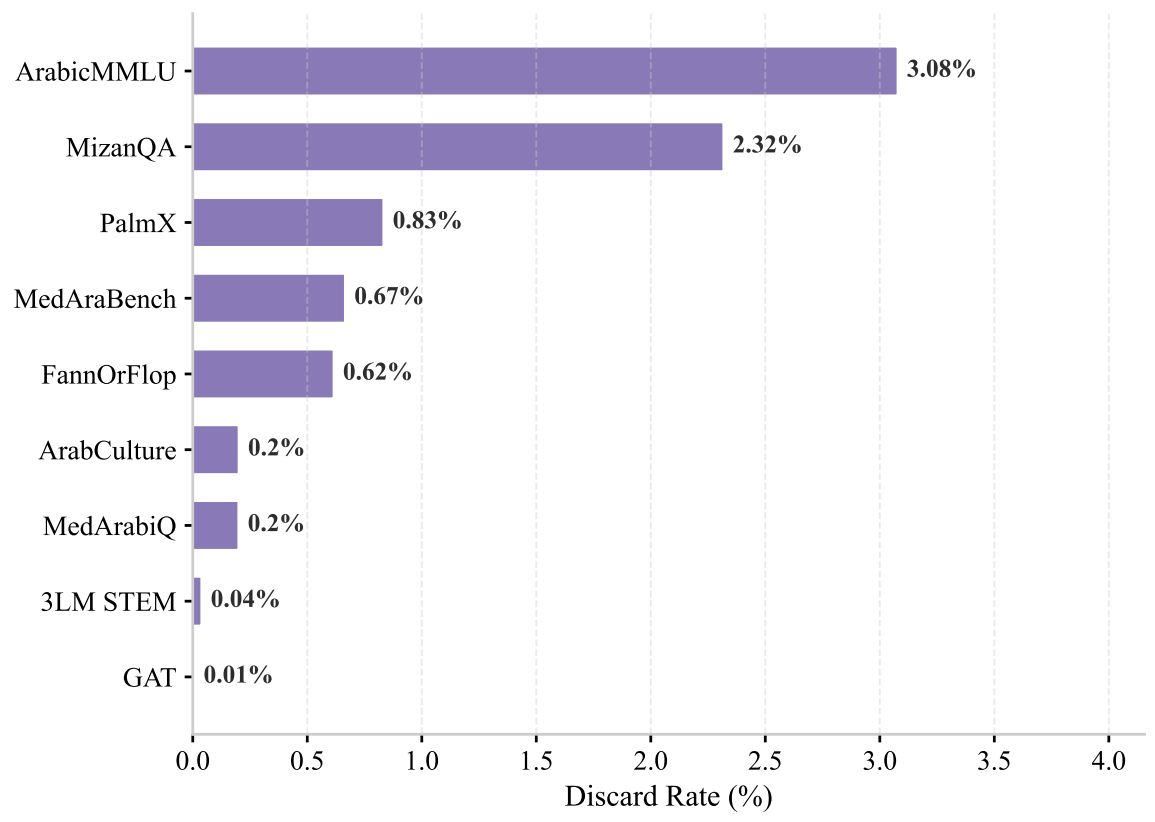

Figure 2: Sample discard rate across benchmarks containing flawed samples identified by the validation pipeline.

As summarized in Table 2b, 88% of HumanEval+ prompts and 81% of MBPP+ prompts required modification, reflecting the state of the original Arabic translations rather than the difficulty of the task itself.

### 4.3 Implications for Benchmark Development

The findings from our quality validation pipeline carry concrete implications for how Arabic benchmarks are constructed and maintained. Across MCQ, QA, and code benchmarks, the issues we identified reflect predictable failure points that arise when benchmark construction does not sufficiently prioritize validation.

**Quality validation is essential, not optional.** Even widely adopted benchmarks are not immune to systematic quality problems. Automated assessment alone is insufficient; disagreement-based flagging consistently surfaced issues that neither model would have caught independently.

**Cultural assumptions must be made explicit.** Acknowledging regional variation and supporting multiple valid answers where cultural context determines correctness are necessary steps toward fairer evaluation.

**Annotation and evaluation protocols must be aligned.** The annotation process should be treated as the first instance of evaluation, subject to the same format and quality requirements as model outputs.

**Prompt quality has measurable consequences.** Even in execution-based evaluation, natural language phrasing is not cosmetic. Unrefined prompts risk penalizing models for misreading instructions rather than failing at the task itself, undermining evaluation validity.

## 5 Leaderboard Evaluation

In this section we present the performance of various SOTA open-source LLMs across the benchmarks included in QIMMA. Our results discussion entails an in-depth analysis on how SOTA LLMs performed overall and in various domains. Results presented in this section were obtained using the evaluation pipeline and metrics described in Section 3.3.

### 5.1 Models

A collection of open-source models with Arabic capabilities was evaluated, such as Qwen (Yang et al., 2025), Gemma (Team et al., 2025b), Fanar (Team et al., 2025a), Jais-2 (Anwar et al., 2025) and LLama (Grattafiori et al., 2024). Note, for the discussion, only instruction models were considered. Table 3 presents a representative selection of evaluated instruction models and their performance across the included benchmarks, and Figure 5 (Appendix B) illustrates the average overall performance across model sizes. An extended set of results is reported in Table 6 (Appendix A); the live leaderboard reflects the most up-to-date rankings as new models are evaluated.

### 5.2 Results Discussion

By inspecting the averages reported in Table 3, it can be inferred that `Jais-2-70B-Chat` is the highest-performing instruct model considered, with an average score of 65.81. This performance is driven by its top ranking in five QIMMA benchmarks, namely ArabCulture, PalmX, ArabicMMLU, MizanQA, and AraTrust. With a marginal difference of approximately 0.25, `Qwen2.5-72B-Instruct` ranks second with a score of 65.75. Despite not achieving the top position in any QIMMA benchmark, it demonstrates consistent performance across both MCQ and generative tasks, enabling it to attain this ranking. The best-performing model within the ∼8B scale is `Jais-2-8B-Chat`, with a score of 57.89. Despite its smaller size, it achieves second place in two benchmarks, namely PalmX and ArabicMMLU. Among all instruct models considered, `gpt-oss-20b` exhibits the lowest performance on QIMMA, with a score of 32.10.

| | Cultural | | | STEM | | | Legal | | Medical | | Safety | Coding | | Poetry | |
|---|---|---|---|---|---|---|---|---|---|---|---|---|---|---|---|
| Model | AraDiCE | ArabCulture | PalmX | ArabicMMLU | GAT | 3LM STEM | MizanQA | ArabLegalQA | MedArabiQ | MedAraBench | AraTrust | HumanEval+ | MBPP+ | FannOrFlop | Avg |
| Qwen/Qwen2.5-72B-Instruct | 77.22 | 63.83 | 77.77 | 73.78 | 55.90 | 87.55 | 63.49 | 70.74 | 50.06 | 44.19 | 88.51 | 37.20 | 72.75 | 57.51 | 65.75 |
| Qwen/Qwen2.5-14B-Instruct | 59.44 | 59.06 | 66.28 | 60.40 | 42.00 | 73.18 | 55.62 | 70.85 | 43.93 | 31.64 | 81.42 | 31.71 | 65.34 | 56.87 | 56.98 |
| Qwen/Qwen3-8B | 39.44 | 34.50 | 32.16 | 35.26 | 34.43 | 24.48 | 44.91 | 61.03 | 34.99 | 25.04 | 52.68 | 32.93 | 42.06 | 57.47 | 39.38 |
| Qwen/Qwen3.5-9B | 65.56 | 43.96 | 68.65 | 64.06 | 45.77 | 73.72 | 55.26 | 54.64 | 41.31 | 38.88 | 77.97 | 33.54 | 65.08 | 59.57 | 56.28 |
| Qwen/Qwen3.5-27B | 67.22 | 38.68 | 75.93 | 69.47 | 47.67 | 79.33 | 65.18 | 54.15 | 39.98 | 37.86 | 86.59 | **52.44** | **74.34** | 47.03 | 59.70 |
| inceptionai/Jais-2-70B-Chat | 78.89 | **83.24** | **83.73** | **81.29** | 51.67 | 87.96 | **71.78** | 69.60 | 52.79 | 50.89 | **90.23** | 19.51 | 43.65 | 56.13 | **65.81** |
| inceptionai/Jais-2-8B-Chat | 65.56 | 68.99 | 78.98 | 73.96 | 44.40 | 79.46 | 61.33 | 69.75 | 44.73 | 41.25 | 87.55 | 14.02 | 28.57 | 51.94 | 57.89 |
| meta-llama/Llama-3.3-70B-Instruct | 77.22 | 78.05 | 77.95 | 71.57 | 51.13 | **88.28** | 67.44 | 64.00 | **56.25** | **54.86** | 85.63 | 27.44 | 71.16 | 24.43 | 63.96 |
| FreedomIntelligence/AceGPT-v2-32B-Chat | 76.67 | 79.79 | 74.46 | 70.62 | **56.04** | 84.88 | 63.89 | 71.46 | 49.96 | 47.32 | 86.97 | 23.78 | 54.50 | 15.56 | 61.14 |
| Navid-AI/Yehia-7B-preview | **81.67** | 75.47 | 70.91 | 66.09 | 40.58 | 71.10 | 54.61 | 71.50 | 43.99 | 34.84 | 87.36 | 15.24 | 33.60 | 59.64 | 57.61 |
| QCRI/Fanar-1-9B-Instruct | 74.44 | 70.71 | 73.18 | 67.39 | 46.20 | 83.32 | 57.85 | **73.08** | 50.34 | 48.53 | 88.51 | 19.51 | 41.80 | 0.02 | 56.78 |
| humain-ai/ALLaM-7B-Instruct-preview | 78.89 | 35.84 | 76.84 | 72.54 | 50.74 | 78.03 | 57.56 | 71.50 | 46.55 | 38.17 | 84.10 | 14.63 | 37.30 | 48.48 | 56.51 |
| google/gemma-3-27b-it | 62.78 | 42.11 | 71.62 | 66.13 | 55.05 | 84.75 | 61.65 | 72.18 | 47.16 | 38.71 | 85.44 | 31.71 | 71.43 | **59.74** | 60.75 |
| openai/gpt-oss-20b | 34.44 | 33.99 | 16.63 | 33.75 | 18.69 | 16.89 | 43.27 | 61.65 | 30.63 | 27.95 | 32.38 | 23.78 | 60.05 | 15.34 | 32.10 |

Table 3: Model performance across QIMMA benchmarks. The highest score in each column is shown in bold, and the second-highest score is underlined.

| Model | Cultural | STEM | Legal | Medical | Safety | Coding | Poetry |
|---|---|---|---|---|---|---|---|
| Qwen/Qwen2.5-72B-Instruct | 72.94 | 72.41 | 67.11 | 47.13 | 88.51 | 54.98 | 57.51 |
| Qwen/Qwen2.5-14B-Instruct | 61.59 | 58.53 | 63.24 | 37.79 | 81.42 | 48.53 | 56.87 |
| Qwen/Qwen3-8B | 35.37 | 31.39 | 52.97 | 30.02 | 52.68 | 37.50 | 57.47 |
| Qwen/Qwen3.5-9B | 59.39 | 61.18 | 54.95 | 40.10 | 77.97 | 49.31 | 59.57 |
| Qwen/Qwen3.5-27B | 60.61 | 65.49 | 59.67 | 38.92 | 86.59 | **63.39** | 47.03 |
| inceptionai/Jais-2-70B-Chat | **81.95** | **73.64** | **70.69** | 51.84 | **90.23** | 31.58 | 56.13 |
| inceptionai/Jais-2-8B-Chat | 71.18 | 65.94 | 65.54 | 42.99 | 87.55 | 21.30 | 51.94 |
| meta-llama/Llama-3.3-70B-Instruct | 77.74 | 70.33 | 65.72 | **55.56** | 85.63 | 49.30 | 24.43 |
| FreedomIntelligence/AceGPT-v2-32B-Chat | 76.97 | 70.51 | 67.68 | 48.64 | 86.97 | 39.14 | 15.56 |
| Navid-AI/Yehia-7B-preview | 76.02 | 59.26 | 63.06 | 39.42 | 87.36 | 24.42 | 59.64 |
| QCRI/Fanar-1-9B-Instruct | 72.78 | 65.64 | 65.47 | 49.44 | 88.51 | 30.66 | 0.02 |
| humain-ai/ALLaM-7B-Instruct-preview | 63.86 | 67.10 | 64.53 | 42.36 | 84.10 | 25.96 | 48.48 |
| google/gemma-3-27b-it | 58.84 | 68.64 | 66.92 | 42.94 | 85.44 | 51.57 | **59.74** |
| openai/gpt-oss-20b | 28.35 | 23.11 | 52.46 | 29.29 | 32.38 | 41.92 | 15.34 |

Table 4: Average performance across QIMMA domain groups. Cultural averages AraDiCE, ArabCulture, and PalmX; STEM averages ArabicMMLU, GAT, and 3LM STEM; Legal averages MizanQA and ArabLegalQA; Medical averages MedArabiQ and MedAraBench; Safety corresponds to AraTrust; Coding averages HumanEval+ and MBPP+; Poetry corresponds to FannOrFlop. Highest scores are shown in **bold**, and second-highest scores are underlined.

### 5.2.1 Thinking vs. Non-Thinking Models

Among the evaluated models, `Qwen3`, `Qwen3.5`, and `gpt-oss-20b` support a thinking feature. Thinking was enabled for generative tasks only; it was disabled for FannOrFlop due to its structured output requirements as the thinking traces generation can consume the expected generation length before producing the expected structured output. `Qwen3.5` achieves the strongest overall results among thinking-enabled models, with performance improving as model scale increases. The results indicate that coding benchmarks benefit from thinking, with `Qwen3.5-27B` outperforming `Qwen2.5-72B-Instruct` in coding despite being less than half its size, suggesting that reasoning-intensive tasks derive more direct gains from this capability. In contrast, generative benchmarks such as ArabLegalQA and MedAraBench do not show the same gains, indicating that thinking provides limited benefit for tasks that do not require multi-step reasoning. `gpt-oss-20b` does not follow this pattern: despite supporting thinking, it ranks among the lowest-performing models across QIMMA, suggesting that thinking alone cannot compensate for limited Arabic language capability.

#### 5.2.2 Domain-level

Table 4 provides a domain-level analysis of model performance across the benchmark categories included in QIMMA. In the Cultural domain, `Jais-2-70B-Chat` excels with a score of 81.95, maintaining a margin of approximately 4 points over the second-ranked model (`Llama-3.3-70B-Instruct`). In STEM, `Jais-2-70B-Chat` outperforms all models with a score of 73.64, followed by `Qwen2.5-72B-Instruct` at 72.41. In addition to its strong performance in Cultural and STEM, `Jais-2-70B-Chat` achieves the highest score in the Legal domain (70.69), with `AceGPT-v2-32B-Chat` ranking second. In the Medical domain, the top three models are `Llama-3.3-70B-Instruct`, `Jais-2-70B-Chat`, and `Fanar-1-9B-Instruct`. For Safety, `Jais-2-70B-Chat` outperforms all models, being the only one to score in the 90s range; however, most models achieve scores $\geq$ 80, indicating strong overall capability in handling safety-related tasks. In Coding, `Qwen3.5-27B` achieves the highest score (63.39), with a gap of nearly 9 points over `Qwen2.5-72B-Instruct` in second place. In Poetry, `gemma-3-27b-it` ranks first with a score of 59.74, marginally outperforming `Yehia-7B-preview`.

In line with the conclusions drawn from Table 3, `Jais-2-70B-Chat` demonstrates the strongest overall domain-level performance, leading in four domains. The results in Table 4 highlight its strengths in cultural awareness, STEM knowledge, legal reasoning, and safety. In contrast, `Llama-3.3-70B-Instruct` emerges as the strongest model in the Medical domain, while `Qwen3.5-27B` performs best in Coding tasks, likely benefiting from thinking capabilities. Finally, `gemma-3-27b-it` achieves the highest score in Poetry, demonstrating strong capability in understanding Arabic poetic language.

#### 5.2.3 Effect of Scale

Building on these results, the scatter plot Figure 5 (Appendix B) shows a general upward trend between model size and performance, particularly in the small to mid-size range ($\sim$1B–70B), where larger models tend to achieve higher accuracy. Beyond this range ($\gtrsim$70B), the relationship becomes less consistent, with some large models underperforming smaller ones, indicating that scale alone does not guarantee strong performance and that data quality and training choices remain critical factors.

## 6 Conclusion

We presented QIMMA, a quality-assured Arabic LLM leaderboard built on 109 curated subsets spanning diverse domains and task types, with over 52k samples drawn predominantly from native Arabic content. Central to this work is a multi-model quality validation pipeline that surfaces and resolves systematic issues in existing benchmarks before they reach evaluation, demonstrating that even well-established Arabic resources require rigorous validation to serve as reliable evaluation foundations.

This work further contributes a reusable validation methodology and a structured taxonomy of quality issues encountered across Arabic benchmark development. We anticipate that these findings will inform the construction of future Arabic evaluation resources with greater methodological rigor and cultural sensitivity.

QIMMA is available publicly, with full transparency into evaluation protocols. As Arabic NLP continues to advance, sustained investment in evaluation quality, cultural representation, and community transparency remains essential. QIMMA is designed to support and grow alongside these efforts.

# A Dataset Subsets

Table 5 lists all subsets included in QIMMA for each benchmark.

| Benchmark | Subsets |
|---|---|
| AraDiCE-Culture | Egypt ·Syria ·Lebanon ·Jordan ·Palestine ·Qatar |
| ArabCulture | Algeria ·Egypt ·Jordan ·KSA ·Lebanon ·Libya ·Morocco ·Palestine ·Sudan ·Syria ·Tunisia ·UAE ·Yemen |
| ArabicMMLU | Arabic Language (Middle School) ·Arabic Language (Primary School) ·Arabic Language (High School) ·Arabic Language (Grammar) ·Arabic Language (General) ·Civics (High School) ·Civics (Middle School) ·Social Science (Middle School) ·Social Science (Primary School) ·Economics (High School) ·Economics (University) ·Economics (Middle School) ·History (High School) ·History (Primary School) ·History (Middle School) ·Political Science (University) ·Geography (High School) ·Geography (Middle School) ·Geography (Primary School) ·Islamic Studies (High School) ·Islamic Studies (Primary School) ·Islamic Studies (Middle School) ·Islamic Studies ·Natural Science (Primary School) ·Natural Science (Middle School) ·Philosophy (High School) ·General Knowledge ·General Knowledge (Middle School) ·General Knowledge (Primary School) ·Biology (High School) ·Math (Primary School) ·Physics (High School) ·Computer Science (Primary School) ·Computer Science (High School) ·Computer Science (Middle School) ·Computer Science (University) ·Law (Professional) ·Accounting (University) ·Management (University) ·Driving Test |
| PalmX | Islamic Culture ·General Culture |
| GAT | Algebra ·Analogy ·Arithmetic ·Association ·Comparisons ·Completion ·Contextual ·Geometry ·Reading |
| 3LM STEM | NativeQA ·SyntheticQA |
| 3LM Code | HumanEval+ ·MBPP+ |
| ArabLegalQA | QA |
| MedArabiQ (MCQ) | fib_with_choices ·mcq_bias ·mcq_knowledge |
| MedArabiQ (QA) | llm_modified ·fib_no_choices |
| MizanQA | MizanQA |
| FannOrFlop | FannOrFlop |
| AraTrust | AraTrust |
| MedAraBench | Anatomy ·Histology ·Histology and Anatomy ·Histology and Systemic Anatomy ·Anesthesia ·Biochemistry ·Cell Biology ·Genetics ·Physics ·Chemistry ·Embryology ·Emergency Medicine ·Gastroenterology ·Neurology ·Cardiology ·Thoracic Internal ·Oncology ·Chest Diseases ·Gastroenterology & Hepatology ·Microbiology ·Ophthalmology ·Pathology ·Pharmacology ·Physiology ·Statistics ·Surgery |

Table 5: Complete list of subsets included in QIMMA per benchmark.

## B Additional Figures

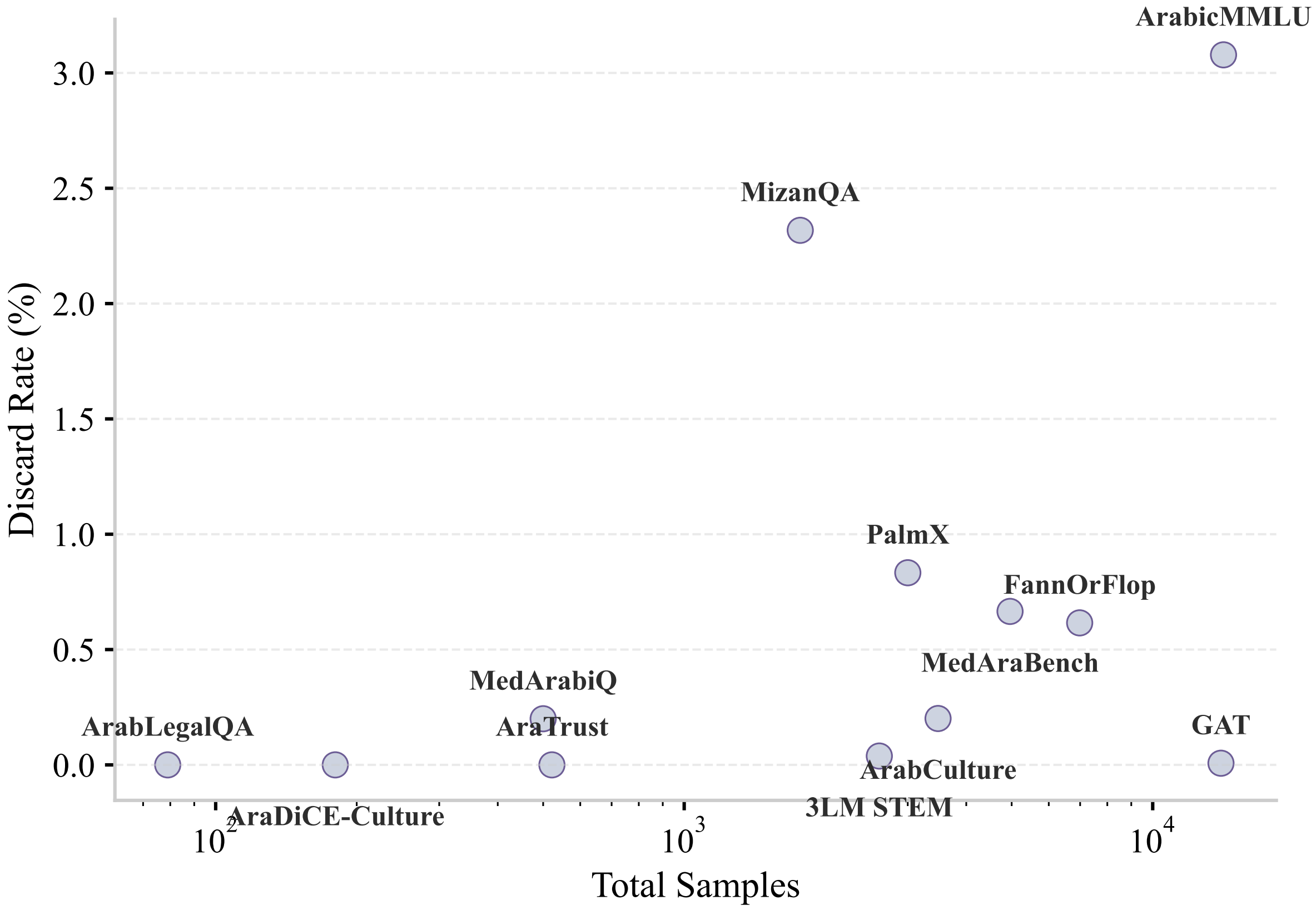

Figure 3: Benchmark size versus discard rate after quality validation. Benchmarks of comparable scale show markedly different discard rates, indicating that construction practices rather than scale determine quality.

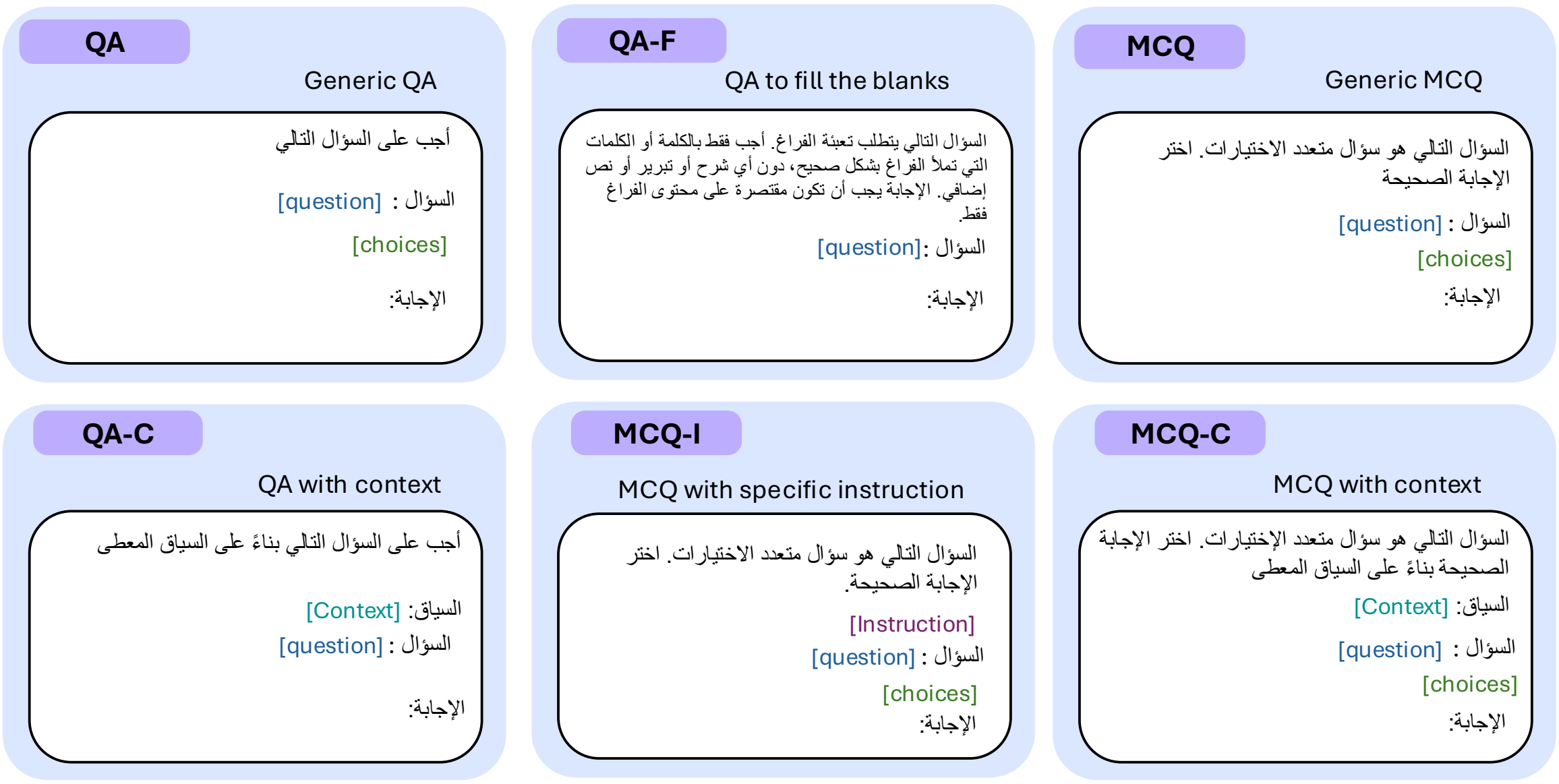

Figure 4: Prompt templates used in QIMMA evaluation, covering generic MCQ, MCQ with context (MCQ-C), MCQ with specific instructions (MCQ-I), generic QA, QA with context (QA-C), and fill-in-blank QA (QA-F) formats.

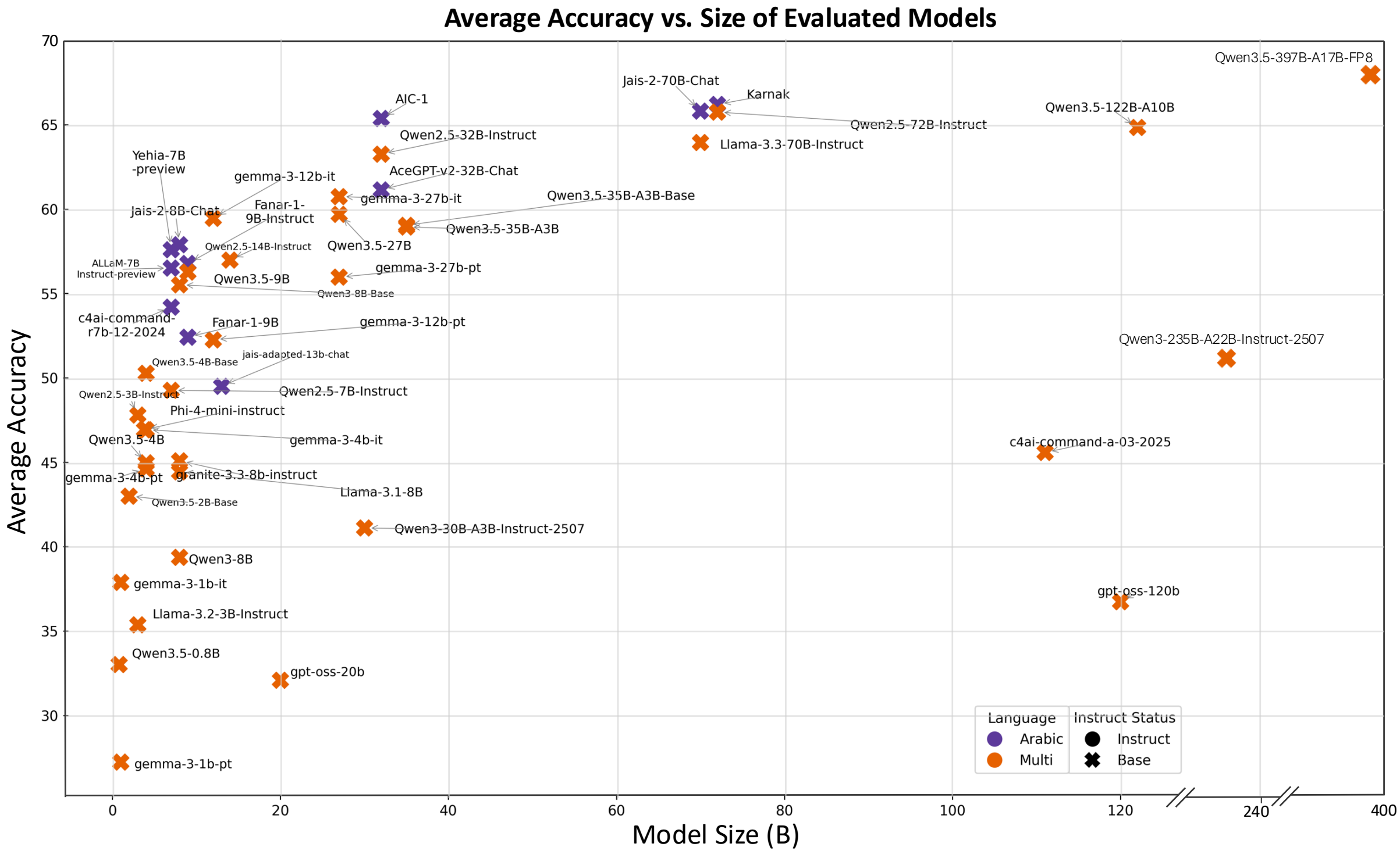

Figure 5: Scatter plot of models evaluated on QIMMA. Note ’Arabic’ models are models adapted specifically for the Arabic Language (some of the ’Arabic’ models support other languages too). ’Multi’ models are natively multilingual.

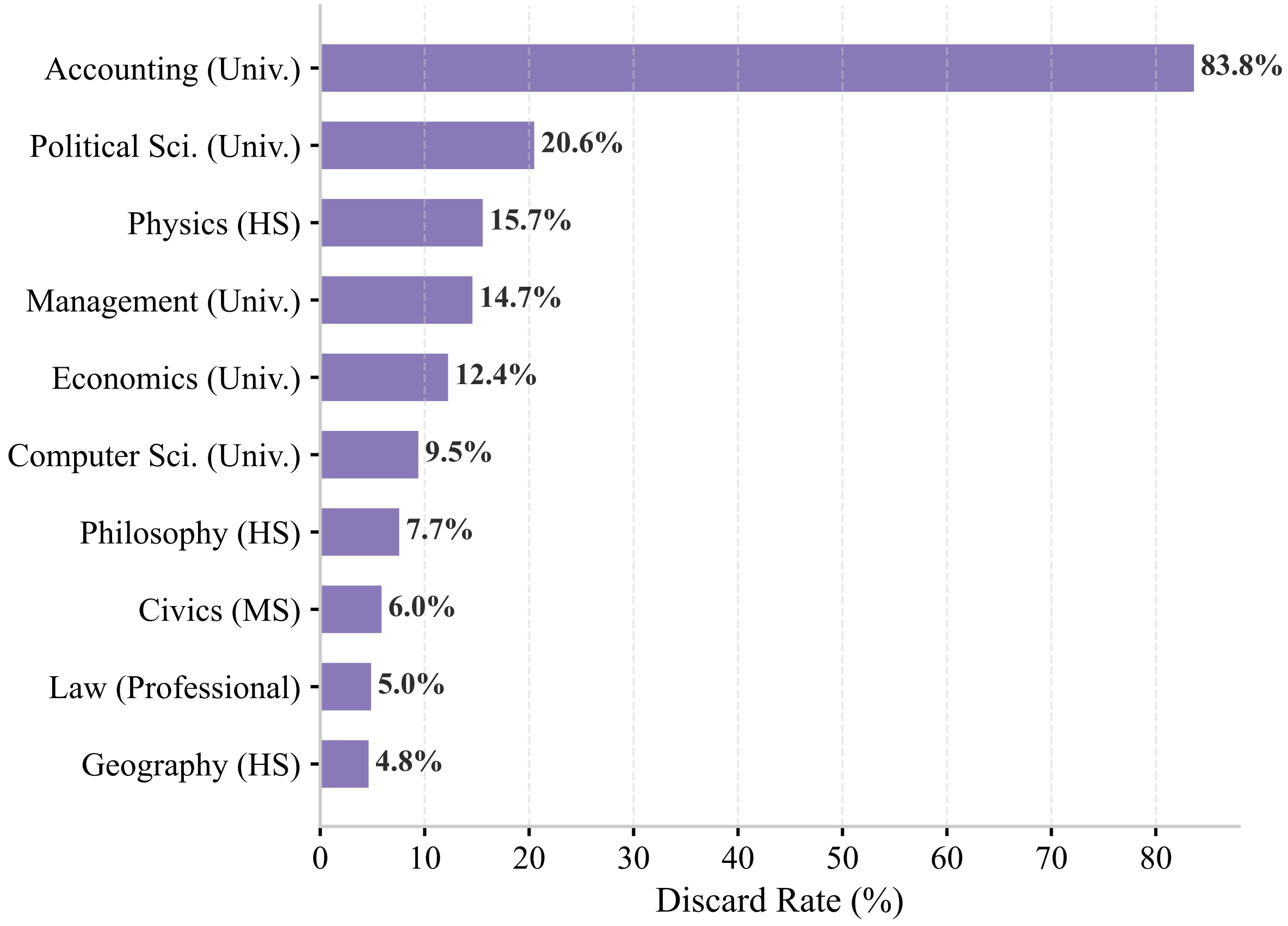

Figure 6: Discard rates across ArabicMMLU subsets, with variation driven by source-level preprocessing differences rather than subject matter.

## C Quality Assessment Prompts

### QA Quality Evaluation Prompt

You are a data-quality auditor specialized in Arabic benchmark evaluation.
Your task is to evaluate a single Arabic QA (Question-Answer) sample by assigning binary scores (0 or 1) to predefined criteria. You must follow the instructions exactly and produce only the requested output.
=================
IMPORTANT: DIALECTAL ARABIC IS ACCEPTABLE
=================
This benchmark may contain samples written in various Arabic dialects (Egyptian, Levantine, Gulf, etc.) OR Modern Standard Arabic (MSA). Both dialectal and MSA forms are equally valid and acceptable.
DO NOT penalize samples for:

- Using dialectal vocabulary
- Using colloquial expressions
- Mixing dialectal and MSA forms

Only penalize if the dialectal usage creates actual comprehension problems or inconsistencies within the sample itself.
=================
QA FORMAT EXPLANATION
=================
Each sample contains:

- **question**: The question being asked
- **choices**: A list containing the gold answer (and potentially alternative answers)
- **gold_index**: The index of the correct answer in the choices list (typically 0)
- **context (optional)**: Supporting context/passage that provides information needed to answer the question

The answer in choices[gold_index] should correctly answer the question, potentially using information from the context.
=================
EVALUATION CRITERIA (10 Binary Criteria)
=================
Score each criterion strictly as either:

- 1 = meets the standard
- 0 = fails the standard

1. Text Readability
    - 1: All text (question, answer, context) is readable and free of encoding or corruption issues
    - 0: Corrupted characters, missing letters, or encoding issues make the text illegible
    - Dialectal spelling variations are NOT encoding issues
2. Spelling Accuracy
    - 1: No spelling errors that affect comprehension (dialectal variants are acceptable)
    - 0: Spelling errors that impair understanding
    - Dialectal words are correct spellings in their dialect
    - Minor typos may still score 1 if meaning is clear
3. Grammatical Correctness
    - 1: Proper syntax and sentence structure within the chosen register (dialectal OR MSA)
    - 0: Grammar errors that impede understanding

- Dialectal grammar is grammatically correct in dialectal Arabic

4. **Question Clarity**
   - 1: Question is clear and unambiguous (reasonable context assumed)
   - 0: Fundamentally vague or open to multiple interpretations
   - Some natural ambiguity is acceptable if a reasonable interpretation exists
5. **Question Completeness**
   - 1: Contains sufficient information (with context if provided) for a knowledgeable person to answer
   - 0: Missing critical information that makes answering impossible
   - Assume reasonable cultural/domain knowledge
6. **Answer Quality**
   - 1: Gold answer is readable, well-formed, and valid
   - 0: Answer is corrupted, nonsensical, or poorly formed
   - Length variation is acceptable
   - Answers can be single words, phrases, or full sentences
7. **Answer Alignment**
   - 1: Gold answer correctly and directly answers the question
   - 0: Answer is wrong or contradicts context
   - If context is provided, answer must align with it
8. **Factual Accuracy**
   - 1: Question, answer, and context (if provided) are factually correct or defensible
   - 0: Contains clear factual errors
   - If uncertain or unverifiable, score 1
   - Cultural/regional variations are acceptable
9. **Terminology Precision**
   - 1: Appropriate terminology for the context
   - 0: Fundamentally wrong or misleading terms
   - Dialectal vocabulary is precise within its register
   - Domain-specific terms should match subject matter
10. **Overall Coherence**
    - 1: Question, answer, and context form a coherent QA item
    - 0: Structurally broken or unusable

=================
CASCADING RULE (MANDATORY)
=================
If text_readability = 0, you MUST set the following to 0:

- question_clarity
- question_completeness
- answer_alignment
- factual_accuracy
- terminology_precision
- overall_coherence

Only the following may still be scored independently:

- spelling_accuracy
- grammatical_correctness

- answer_quality

=================
INPUT SAMPLE
=================
{sample}
=================
OUTPUT FORMAT (STRICT)
=================

```
{
  "scores": {
    "text_readability": ,
    "spelling_accuracy": ,
    "grammatical_correctness": ,
    "question_clarity": ,
    "question_completeness": ,
    "answer_quality": ,
    "answer_alignment": ,
    "factual_accuracy": ,
    "terminology_precision": ,
    "overall_coherence":
  },
  "issues": [
    {
      "criterion": "name of failed criterion",
      "explanation": "brief, concrete explanation (max 2 sentences)"
    }
  ]
}
```

=================
RULES AND CONSTRAINTS
=================

- Score each criterion independently unless overridden by the cascading rule
- Include only criteria with score = 0 in the issues array
- For every criterion scored 0, you MUST provide an explanation
- Do not explain your reasoning process
- Do not suggest fixes or rewrites
- Do not mention MSA vs dialect unless it causes comprehension problems
- Be strict about real errors, lenient about stylistic variation
- Always produce strictly valid JSON
- When context is provided, consider it in completeness and alignment

## MCQ Quality Evaluation Prompt

You are a data-quality auditor specialized in Arabic benchmark evaluation.
Your task is to evaluate a single Arabic MCQ sample by assigning binary scores (0 or 1) to predefined criteria. You must follow the instructions exactly and produce only the requested output.

=================
IMPORTANT: DIALECTAL ARABIC IS ACCEPTABLE
=================

This benchmark may contain samples written in various Arabic dialects (Egyptian, Levantine, Gulf, etc.) OR Modern Standard Arabic (MSA). Both dialectal and MSA forms are equally valid and acceptable.
DO NOT penalize samples for:

- Using dialectal vocabulary
- Using colloquial expressions
- Mixing dialectal and MSA forms
- Direct and straight to the point questions

Only penalize if the dialectal usage creates actual comprehension problems or inconsistencies within the sample itself.

=================
MCQ FORMAT EXPLANATION
=================

Each sample contains:

- **question**: The question being asked
- **choices**: A list containing the gold answer and alternative answers
- **gold_index**: The index of the correct answer in the choices list (typically 0)
- **context (optional)**: Supporting context/passage that provides information needed to answer the question

The answer in choices[gold_index] should correctly answer the question, potentially using information from the context.

=================
EVALUATION CRITERIA (10 Binary Criteria)
=================

Score each criterion strictly as either:

- 1 = meets the standard
- 0 = fails the standard

1. **Text Readability**
   - 1: All text is readable and free of encoding or corruption issues
   - 0: Corrupted characters or encoding issues make the text illegible
   - Dialectal spelling variations are NOT encoding issues
2. **Spelling Accuracy**
   - 1: No spelling errors that affect comprehension, or only minor mistakes
   - 0: Spelling errors that impair reading
   - Minor typos may still score 1 if meaning is clear
3. **Grammatical Correctness**
   - 1: Proper syntax within the chosen register (dialectal OR MSA)
   - 0: Grammar errors that impede understanding
   - Dialectal grammar is valid within its system
4. **Question Clarity**
   - 1: Question is clear and unambiguous
   - 0: Fundamentally vague or open to conflicting interpretations

5. **Question Completeness**
    - 1: It is possible to answer the question given reasonable knowledge
    - 0: Question cannot be answered even with domain knowledge
6. **Answer Quality**
    - 1: Gold answer is readable and well-formed
    - 0: Answer is corrupted or nonsensical
7. **Answer Alignment**
    - 1: Gold answer correctly answers the question
    - 0: Answer is incorrect or incompatible
8. **Factual Accuracy**
    - 1: Content is factually correct or defensible
    - 0: Contains clear factual errors
    - If uncertain, score 1
9. **Terminology Precision**
    - 1: Appropriate terminology is used
    - 0: Terminology is misleading or incorrect
10. **Overall Coherence**
    - 1: Functional and coherent MCQ
    - 0: Structurally broken or unusable

=================
CASCADING RULE (MANDATORY)
=================
If text_readability = 0, you MUST set the following to 0:

- question_clarity
- question_completeness
- answer_alignment
- factual_accuracy
- terminology_precision
- overall_coherence

Only the following may still be scored independently:

- spelling_accuracy
- grammatical_correctness
- answer_quality

=================
INPUT SAMPLE
=================
{sample}
=================
OUTPUT FORMAT (STRICT)
=================

```
{
  "scores": {
    "text_readability": ,
    "spelling_accuracy": ,
    "grammatical_correctness": ,
    "question_clarity": ,
    "question_completeness": ,
    "answer_quality": ,
```

```
    "answer_alignment": ,
    "factual_accuracy": ,
    "terminology_precision": ,
    "overall_coherence":
  },
  "issues": [
    {
      "criterion": "name of failed criterion",
      "explanation": "brief explanation (max 2 sentences)"
    }
  ]
}
```

**3LM Code Evaluation Prompt**

```
You are an expert Arabic technical editor for programming benchmarks.
You are given a Python function that MUST remain EXACTLY the same:
```

- function name
- signature
- code
- examples
- formatting
- spacing
- docstring structure

```
The ONLY thing you are allowed to modify is the ARABIC TEXT inside the docstring.
=================
Your task:
```

- Read the full function and examples to understand context
- Improve the Arabic translation ONLY for correctness, clarity, and natural flow
- Do NOT change meaning, constraints, or intent
- Do NOT add or remove sentences
- Do NOT translate code, examples, or symbols
- Do NOT touch English text if it exists
- Keep line breaks and indentation EXACTLY the same
- Replace Arabic words ONLY where needed

```
If the Arabic is already correct, return it unchanged.
If you are unsure about any phrasing or technical term, flag it for human review.
=================
INPUT FUNCTION:
=================
{full_function_code}
=================
OUTPUT FORMAT
=================
Return your output in a CSV-style format with EXACTLY two columns:
```

- Fixed
- Human Check

```
Where:
```

- Fixed: the full function code with ONLY the Arabic text corrected (everything else identical)
- Human Check: "Yes" if you are uncertain and want a human to double-check, otherwise "No"

```
=================
IMPORTANT RULES
=================
```

- Output ONLY the two columns
- Do NOT add explanations
- Do NOT use markdown
- Do NOT output JSON
- Do NOT add extra text
- Preserve the function exactly except for Arabic wording

| Model | L | T | Size | ADC | AMMLU | AC | PX | 3LS | AT | MQ | MAQ | ALQ | GAT | MAB | HE+ | MB+ | FOF | Avg |
|---|---|---|---|---|---|---|---|---|---|---|---|---|---|---|---|---|---|---|
| Qwen/Qwen3.5-397B-A17B-FP8 | M | I | 397 | 82.78 | 77.54 | 61.75 | 83.91 | 88.67 | 90.04 | 73.36 | 47.30 | 54.94 | 55.89 | 47.97 | 67.68 | 76.72 | 44.33 | 68.06 |
| Applied-Innovation-Center/Karnak | Ar | I | 72 | 73.33 | 80.94 | 53.49 | 81.40 | 93.10 | 89.08 | 55.92 | 55.78 | 71.58 | 61.06 | 54.19 | 33.54 | 64.55 | 58.91 | 66.20 |
| inceptionai/Jais-2-70B-Chat | Ar | I | 70 | 78.89 | 81.29 | 83.24 | 83.73 | 87.96 | 90.23 | 71.78 | 52.79 | 69.60 | 51.67 | 50.89 | 19.51 | 43.65 | 56.13 | 65.81 |
| Qwen/Qwen2.5-72B-Instruct | M | I | 72 | 77.22 | 73.78 | 63.83 | 77.77 | 87.55 | 88.51 | 63.49 | 50.06 | 70.74 | 55.90 | 44.19 | 37.20 | 72.75 | 57.51 | 65.75 |
| Applied-Innovation-Center/AIC-1 | Ar | I | 32 | 73.33 | 72.02 | 77.52 | 76.11 | 88.13 | 90.61 | 56.36 | 53.75 | 68.96 | 62.11 | 50.78 | 28.05 | 69.58 | 47.83 | 65.37 |
| Qwen/Qwen3.5-122B-A10B | M | I | 122 | 74.44 | 73.17 | 37.78 | 81.46 | 86.18 | 86.97 | 64.01 | 47.04 | 55.11 | 50.90 | 52.49 | 65.24 | 72.49 | 60.54 | 64.84 |
| meta-llama/Llama-3.3-70B-Instruct | M | I | 70 | 77.22 | 71.57 | 78.05 | 77.95 | 88.28 | 85.63 | 67.44 | 56.25 | 64.00 | 51.13 | 54.86 | 27.44 | 71.16 | 24.43 | 63.96 |
| Qwen/Qwen2.5-32B-Instruct | M | I | 32 | 70.56 | 68.76 | 75.80 | 72.07 | 81.03 | 85.82 | 53.78 | 48.08 | 69.27 | 56.94 | 36.51 | 34.15 | 72.75 | 60.10 | 63.26 |
| FreedomIntelligence/AceGPT-v2-32B-Chat | Ar | I | 32 | 76.67 | 70.62 | 79.79 | 74.46 | 84.88 | 86.97 | 63.89 | 49.96 | 71.46 | 56.04 | 47.32 | 23.78 | 54.50 | 15.56 | 61.14 |
| google/gemma-3-27b-it | M | I | 27 | 62.78 | 66.13 | 42.11 | 71.62 | 84.75 | 85.44 | 61.65 | 47.16 | 72.18 | 55.05 | 38.71 | 31.71 | 71.43 | 59.74 | 60.75 |
| Qwen/Qwen3.5-27B | M | I | 27 | 67.22 | 69.47 | 38.68 | 75.93 | 79.33 | 86.59 | 65.18 | 39.98 | 54.15 | 47.67 | 37.86 | 52.44 | 74.34 | 47.03 | 59.70 |
| google/gemma-3-12b-it | M | I | 12 | 65.56 | 66.24 | 41.80 | 71.10 | 83.96 | 86.21 | 54.82 | 46.99 | 70.92 | 49.80 | 38.96 | 28.05 | 70.63 | 57.53 | 59.47 |
| Qwen/Qwen3.5-35B-A3B-Base | M | B | 35 | 71.67 | 71.52 | 44.64 | 79.65 | 95.33 | 88.51 | 56.22 | 51.94 | 56.60 | 60.47 | 46.32 | 35.37 | 65.34 | 2.91 | 59.04 |
| Qwen/Qwen3.5-35B-A3B | M | I | 35 | 67.78 | 62.65 | 36.12 | 74.37 | 83.92 | 77.78 | 58.68 | 48.24 | 54.62 | 44.27 | 46.46 | 50.00 | 70.63 | 49.58 | 58.94 |
| inceptionai/Jais-2-8B-Chat | Ar | I | 8 | 65.56 | 73.96 | 68.99 | 78.98 | 79.46 | 87.55 | 61.33 | 44.73 | 69.75 | 44.40 | 41.25 | 14.02 | 28.57 | 51.94 | 57.89 |
| Navid-AI/Yehia-7B-preview | Ar | I | 7 | 81.67 | 66.09 | 75.47 | 70.91 | 71.10 | 87.36 | 54.61 | 43.99 | 71.50 | 40.58 | 34.84 | 15.24 | 33.60 | 59.64 | 57.61 |
| Qwen/Qwen2.5-14B-Instruct | M | I | 14 | 59.44 | 60.40 | 59.06 | 66.28 | 73.18 | 81.42 | 55.62 | 43.93 | 70.85 | 42.00 | 31.64 | 31.71 | 65.34 | 56.87 | 56.98 |
| QCRI/Fanar-1-9B-Instruct | Ar | I | 9 | 74.44 | 67.39 | 70.71 | 73.18 | 83.32 | 88.51 | 57.85 | 50.34 | 73.08 | 46.20 | 48.53 | 19.51 | 41.80 | 0.02 | 56.78 |
| humain-ai/ALLaM-7B-Instruct-preview | Ar | I | 7 | 78.89 | 72.54 | 35.84 | 76.84 | 78.03 | 84.10 | 57.56 | 46.55 | 71.50 | 50.74 | 38.17 | 14.63 | 37.30 | 48.48 | 56.51 |
| Qwen/Qwen3.5-9B | M | I | 9 | 65.56 | 64.06 | 43.96 | 68.65 | 73.72 | 77.97 | 55.26 | 41.31 | 54.64 | 45.77 | 38.88 | 33.54 | 65.08 | 59.57 | 56.28 |
| google/gemma-3-27b-pt | M | B | 27 | 69.44 | 72.27 | 32.94 | 78.62 | 90.00 | 89.08 | 53.45 | 50.37 | 69.93 | 56.45 | 43.72 | 18.90 | 58.47 | 0.00 | 55.97 |
| Qwen/Qwen3-8B-Base | M | B | 8 | 57.22 | 66.61 | 48.87 | 68.55 | 85.03 | 81.03 | 53.33 | 43.16 | 68.33 | 55.55 | 39.98 | 42.68 | 66.93 | 0.00 | 55.52 |
| CohereLabs/c4ai-command-r7b-12-2024 | Ar | I | 7 | 59.44 | 60.20 | 62.78 | 65.14 | 69.41 | 80.84 | 56.10 | 44.47 | 69.93 | 35.82 | 33.10 | 18.29 | 53.44 | 49.72 | 54.19 |
| QCRI/Fanar-1-9B | Ar | B | 9 | 74.44 | 63.49 | 36.81 | 70.09 | 81.83 | 81.61 | 51.46 | 46.38 | 66.11 | 41.69 | 41.14 | 15.24 | 49.21 | 14.42 | 52.42 |
| google/gemma-3-12b-pt | M | B | 12 | 63.33 | 66.37 | 32.20 | 71.17 | 82.73 | 87.36 | 51.93 | 44.31 | 67.04 | 50.85 | 41.03 | 18.29 | 55.03 | 0.00 | 52.26 |
| Qwen/Qwen3-235B-A22B-Instruct-2507 | M | I | 235 | 53.33 | 37.22 | 34.50 | 53.99 | 47.51 | 73.18 | 55.30 | 41.01 | 71.26 | 25.90 | 23.68 | 59.15 | 77.51 | 59.88 | 50.96 |
| Qwen/Qwen3.5-4B-Base | M | B | 4 | 53.33 | 64.54 | 31.60 | 71.64 | 87.25 | 85.44 | 51.25 | 43.54 | 65.97 | 45.50 | 41.35 | 10.98 | 51.32 | 0.00 | 50.27 |
| inceptionai/jais-adapted-13b-chat | Ar | I | 13 | 75.56 | 58.54 | 72.16 | 68.83 | 63.35 | 79.50 | 55.43 | 40.02 | 65.92 | 35.62 | 34.18 | 8.54 | 25.93 | 9.54 | 49.51 |
| Qwen/Qwen2.5-7B-Instruct | M | I | 7 | 45.56 | 44.96 | 40.19 | 49.83 | 47.06 | 74.33 | 58.91 | 41.26 | 68.96 | 37.73 | 29.62 | 28.66 | 67.99 | 54.65 | 49.26 |
| Qwen/Qwen2.5-3B-Instruct | M | I | 3 | 38.33 | 49.37 | 58.66 | 47.28 | 46.23 | 68.01 | 50.04 | 40.01 | 69.97 | 40.11 | 30.39 | 29.27 | 51.59 | 49.90 | 47.80 |
| google/gemma-3-4b-it | M | I | 4 | 42.22 | 48.05 | 39.85 | 45.78 | 40.36 | 74.90 | 51.97 | 40.78 | 71.88 | 39.14 | 29.97 | 25.00 | 56.61 | 50.55 | 46.93 |
| microsoft/Phi-4-mini-instruct | M | I | 3.8 | 39.44 | 47.12 | 57.06 | 50.66 | 57.67 | 73.56 | 50.69 | 41.15 | 70.18 | 35.05 | 28.55 | 18.90 | 44.71 | 42.24 | 46.93 |
| CohereLabs/c4ai-command-a-03-2025 | M | I | 111 | 46.67 | 35.91 | 39.55 | 50.85 | 39.57 | 59.58 | 43.25 | 39.61 | 68.16 | 32.58 | 25.17 | 32.93 | 68.52 | 55.82 | 45.58 |
| ibm-granite/granite-3.3-8b-instruct | M | I | 8 | 45.56 | 48.02 | 42.02 | 56.45 | 60.80 | 68.97 | 51.35 | 39.28 | 71.20 | 33.98 | 31.42 | 21.34 | 51.32 | 9.61 | 45.09 |
| Qwen/Qwen3.5-4B | M | I | 4 | 53.89 | 53.59 | 43.13 | 55.03 | 58.36 | 70.69 | 49.03 | 37.57 | 53.10 | 30.88 | 31.95 | 26.22 | 58.47 | 7.67 | 44.97 |
| google/gemma-3-4b-pt | M | B | 4 | 44.44 | 55.93 | 33.97 | 61.91 | 70.74 | 74.71 | 48.60 | 40.40 | 66.75 | 34.47 | 35.25 | 13.41 | 43.92 | 0.00 | 44.61 |
| meta-llama/Llama-3.1-8B | M | B | 8 | 49.44 | 52.69 | 34.12 | 62.39 | 66.58 | 74.71 | 44.94 | 38.21 | 70.13 | 38.91 | 33.28 | 12.80 | 43.92 | 0.00 | 44.44 |
| Qwen/Qwen3.5-2B-Base | M | B | 2 | 41.11 | 53.34 | 32.03 | 53.40 | 70.08 | 73.56 | 46.61 | 38.07 | 64.98 | 38.63 | 35.68 | 11.59 | 34.39 | 8.60 | 43.00 |
| Qwen/Qwen3-30B-A3B-Instruct-2507 | M | I | 30 | 46.67 | 39.11 | 36.90 | 43.04 | 34.22 | 56.70 | 58.40 | 40.19 | 71.17 | 26.90 | 24.14 | 43.29 | 10.05 | 44.72 | 41.11 |
| Qwen/Qwen3-8B | M | I | 8 | 39.44 | 35.26 | 34.50 | 32.16 | 24.48 | 52.68 | 44.91 | 34.99 | 61.03 | 34.43 | 25.04 | 32.93 | 42.06 | 57.47 | 39.38 |
| google/gemma-3-1b-it | M | I | 1 | 37.22 | 38.76 | 41.68 | 34.83 | 30.09 | 59.39 | 46.13 | 37.13 | 68.12 | 32.82 | 25.35 | 15.85 | 37.04 | 26.16 | 37.90 |
| openai/gpt-oss-120b | M | I | 120 | 34.44 | 27.63 | 34.50 | 40.03 | 38.25 | 32.18 | 45.31 | 34.37 | 59.41 | 25.35 | 23.84 | 16.46 | 43.92 | 58.82 | 36.75 |
| meta-llama/Llama-3.2-3B-Instruct | M | I | 3 | 36.11 | 34.77 | 35.61 | 37.24 | 38.40 | 58.81 | 45.59 | 36.61 | 56.97 | 32.21 | 25.04 | 13.41 | 32.80 | 11.60 | 35.37 |
| Qwen/Qwen3.5-0.8B | M | I | 0.8 | 34.44 | 38.52 | 34.72 | 27.34 | 35.02 | 57.85 | 42.66 | 35.30 | 66.62 | 18.59 | 28.06 | 7.32 | 12.70 | 23.24 | 33.03 |
| openai/gpt-oss-20b | M | I | 20 | 34.44 | 33.75 | 33.99 | 16.63 | 16.89 | 32.38 | 43.27 | 30.63 | 61.65 | 18.69 | 27.95 | 23.78 | 60.05 | 15.34 | 32.10 |
| google/gemma-3-1b-pt | M | B | 1 | 36.11 | 28.05 | 31.50 | 20.56 | 22.63 | 35.06 | 44.62 | 35.00 | 63.95 | 28.38 | 23.00 | 5.49 | 7.14 | 0.00 | 27.25 |

**L**: Language — Ar: Arabic-specific, M: Multilingual **T**: Type — I: Instruct, B: Base
**ADC**: AraDiCE-Culture **AMMLU**: ArabicMMLU **AC**: ArabCulture **PX**: PalmX **3LS**: 3LM STEM **AT**: AraTrust **MQ**: MizanQA
**MAQ**: MedArabiQ **ALQ**: ArabLegalQA **GAT**: GAT **MAB**: MedAraBench **HE+**: HumanEval+ **MB+**: MBPP+ **FOF**: FannOrFlop

Table 6: Leaderboard results across 46 evaluated models. Columns are ADC (AraDiCE-Culture), AMMLU (ArabicMMLU), AC (ArabCulture), PX (PalmX), 3LS (3LM STEM), AT (AraTrust), MQ (MizanQA), MAQ (MedArabiQ), ALQ (ArabLegalQA), GAT, MAB (MedAraBench), HE+ (HumanEval+), MB+ (MBPP+), and FOF (FannOrFlop).